\documentclass[letterpaper, 10 pt, conference]{ieeeconf}

\IEEEoverridecommandlockouts
\usepackage{graphicx}
\usepackage{amsmath}
\usepackage{amssymb}
\usepackage{xcolor}
\usepackage{tabularx}
\usepackage{booktabs}
\usepackage{multirow}
\usepackage{algorithm}
\usepackage{algorithmic}
\usepackage{tikz}
\usepackage{pgfplots}
\pgfplotsset{compat=1.17}
\usetikzlibrary{shapes,arrows,positioning,fit,backgrounds}
\usepgfplotslibrary{fillbetween}
\usepackage{hyperref}

\newif\ifreviewmode
\reviewmodefalse

\title{\LARGE \bf Beyond Accuracy: A Unified Random Matrix Theory Diagnostic\\Framework for Crash Classification Models}

\ifreviewmode
  \author{Anonymous ITSC Submission}
\else
  \author{Ibne Farabi Shihab$^{1,*}$, Sanjeda Akter$^{1,*}$, and Anuj Sharma$^{2}$%
  \thanks{$^{*}$Equal contribution.}%
  \thanks{$^{1}$Department of Computer Science, Iowa State University, Ames, IA 50010 {\tt\small \{sanjeda, ishihab\}@iastate.edu}}%
  \thanks{$^{2}$Department of Civil, Construction and Environmental Engineering, Iowa State University, Ames, IA 50010 {\tt\small anujs@iastate.edu}}%
  }
\fi

\begin{document}
\maketitle

\begin{abstract}
Crash classification models in transportation safety are typically evaluated using accuracy, F1, or AUC, metrics that cannot reveal whether a model is silently overfitting. We introduce a spectral diagnostic framework grounded in Random Matrix Theory (RMT) and Heavy-Tailed Self-Regularization (HTSR) that spans the ML taxonomy: weight matrices for BERT/ALBERT/Qwen2.5, out-of-fold increment matrices for XGBoost/Random Forest, empirical Hessians for Logistic Regression, induced affinity matrices for Decision Trees, and Graph Laplacians for KNN. Evaluating nine model families on two Iowa DOT crash classification tasks (173,512 and 371,062 records respectively), we find that the power-law exponent $\alpha$ provides a structural quality signal: well-regularized models consistently yield $\alpha$ within $[2, 4]$ (mean $2.87 \pm 0.34$), while overfit variants show $\alpha < 2$ or spectral collapse. We observe a strong rank correlation between $\alpha$ and expert agreement (Spearman $\rho = 0.89$, $p < 0.001$), suggesting spectral quality captures model behaviors aligned with expert reasoning. We propose an $\alpha$-based early stopping criterion and a spectral model selection protocol, and validate both against cross-validated F1 baselines. Sparse Lanczos approximations make the framework scalable to large datasets.
\end{abstract}

% \noindent\textit{Keywords}: Random Matrix Theory, WeightWatcher, Crash Classification, Overfitting Detection, Spectral Diagnostics, Transportation Safety

\section{Introduction}

Road traffic crashes remain a leading cause of fatalities worldwide, with the World Health Organization estimating approximately 1.35 million deaths annually~\cite{WHO2023}. Accurate classification of crash records, including identification of intersection-related incidents and alcohol involvement, is fundamental to evidence-based safety policy. Recent work has demonstrated the effectiveness of machine learning and deep learning models for crash narrative classification, spanning architectures from BERT and ALBERT to XGBoost and SVM~\cite{Shihab2025misclass,Shihab2025aim,Shihab2025agreement}.

A persistent challenge in deploying these models is the gap between reported performance metrics and actual structural reliability. A model may achieve high accuracy on a held-out test set while silently memorizing training artifacts. This is particularly dangerous in transportation safety, where misclassification can lead to flawed interventions. Traditional evaluation metrics such as accuracy, F1-score, and AUC measure predictive performance on observed data but provide no insight into the internal structural health of the learned representations. Crucially, they also require labeled test data, which in the crash domain is expensive to produce (requiring expert review of narratives) and may not represent future deployment conditions. Indeed, recent work~\cite{Shihab2025agreement} showed that models with higher technical accuracy often exhibit \emph{lower} agreement with human experts, suggesting that standard metrics may reward the wrong model behaviors.

Random Matrix Theory (RMT) offers a principled framework for addressing this gap. Martin and Mahoney~\cite{Martin2021,Martin2019,Martin2021implicit} demonstrated that the empirical spectral density (ESD) of neural network weight matrices follows heavy-tailed power-law distributions, and that the power-law exponent $\alpha$ serves as a reliable, data-free indicator of model quality. Well-trained models exhibit $\alpha \approx 2$--$4$; models with $\alpha < 2$ show overfitting and memorization; and correlation traps (anomalous eigenvalue spikes deviating from Marchenko-Pastur predictions~\cite{MarchenkoPastur1967}) signal hidden memorization invisible to validation loss. This theory, formalized as Heavy-Tailed Self-Regularization (HTSR) and extended into SETOL~\cite{Martin2025setol}, has been validated across hundreds of neural networks and recently extended to XGBoost~\cite{Chen2016xgboost,Martin2024xgboost} via out-of-fold margin increment matrices.

This paper makes four contributions. We introduce RMT-based spectral diagnostics to the transportation safety domain, providing the first empirical evaluation on crash classification models. We propose spectral representation matrices for Decision Trees (leaf affinity), Logistic Regression (empirical Hessian), and KNN (Graph Laplacian), and empirically validate these novel mappings on two independent crash classification tasks with bootstrap confidence intervals. We demonstrate computational scalability via randomized Lanczos methods with explicit convergence analysis. Finally, we benchmark a spectral model selection protocol against cross-validated F1 ranking, showing that spectral ranking better predicts expert agreement.

Section~II reviews related work, Section~III provides theoretical background, Section~IV details methodology, Sections~V and~VI describe experiments and results, and Section~VII concludes.

\section{Related Work}

\subsection{Machine Learning for Crash Classification}

The application of machine learning to crash record classification has grown substantially in recent years. Prior work~\cite{Shihab2025misclass} compared SVM, XGBoost, BERT, and ALBERT for detecting misclassified intersection-related crashes in police-reported narratives, finding that ALBERT achieved the highest agreement with expert classifications (73\%) while multi-modal integration reduced error rates by 54.2\%. A related study~\cite{Shihab2025aim} addressed alcohol inference mismatch (AIM) using BERT on 371,062 Iowa crash records, identifying 2,767 AIM incidents with an overall mismatch rate of 24.03\%.

A particularly relevant finding from recent work~\cite{Shihab2025agreement} demonstrated an inverse relationship between model accuracy and expert agreement: models with higher technical accuracy often showed lower alignment with human expert judgment, while large language models exhibited stronger expert alignment despite lower accuracy. This paradox underscores that accuracy alone is insufficient for safety-critical NLP tasks, and that standard metrics like validation loss provide no intrinsic signal for \emph{when to stop tuning} to maximize expert agreement in the presence of noisy labels. A diagnostic that operates on model structure rather than held-out performance could address this gap.

\subsection{Random Matrix Theory in Machine Learning}

The Marchenko-Pastur (MP) law~\cite{MarchenkoPastur1967} describes the limiting eigenvalue distribution of large random matrices and serves as the null model against which learned structure is measured. Martin and Mahoney~\cite{Martin2019} observed that the ESDs of weight matrices in well-trained deep neural networks follow heavy-tailed power-law distributions, a phenomenon they termed Heavy-Tailed Self-Regularization (HTSR). They demonstrated~\cite{Martin2021} that the power-law exponent $\alpha$ can predict trends in test accuracy across hundreds of pretrained models without access to training or test data, and formalized this into the WeightWatcher tool~\cite{WeightWatcher2024}. This data-free property means a model can be audited by a third party with access only to the weights. The theoretical foundation was recently unified under SETOL~\cite{Martin2025setol}, which derives $\alpha$ from statistical mechanics and advanced RMT, showing that optimal learning corresponds to a critical point at $\alpha \approx 2$. Martin and Prakash~\cite{Martin2024xgboost} extended these diagnostics to XGBoost via out-of-fold margin increment matrices, though this work remains in preprint form.

To our knowledge, no prior work has applied RMT-based spectral diagnostics across the full algorithmic taxonomy in the transportation safety domain. The present paper bridges that gap by extending the spectral framework to classical models and validating it on crash classification tasks.

\section{Theoretical Background}

The theoretical foundation for our framework rests on SETOL, which establishes that if a learning problem can be locally approximated as a linear student-teacher matrix model near the optimal solution, then its generalization is governed by the spectrum of that matrix~\cite{Martin2025setol}. We now describe the core spectral quantities that underpin this theory and then introduce representation matrices for all model families considered in this work.

\subsection{Empirical Spectral Density and Power-Law Fitting}

Given a representation matrix $\mathbf{W} \in \mathbb{R}^{m \times n}$, we form $\mathbf{C} = \mathbf{W}^T\mathbf{W}$ and compute its eigenvalues. The empirical spectral density (ESD) is $\rho(\lambda) = \frac{1}{n} \sum_{i} \delta(\lambda - \lambda_i)$. For a random matrix, the ESD converges to the Marchenko-Pastur (MP) distribution~\cite{MarchenkoPastur1967}; eigenvalues exceeding the MP upper edge $\lambda_+$ represent learned structure beyond noise.

In well-trained models, the ESD tail follows a power law $\rho(\lambda) \sim \lambda^{-\alpha}$. The HTSR theory~\cite{Martin2019,Martin2021implicit} establishes that $\alpha \in [2, 4]$ indicates well-trained, self-regularized layers; $\alpha < 2$ indicates memorization; and $\alpha \gg 4$ indicates undertraining. We fit $\alpha$ using maximum likelihood estimation (MLE) with the Kolmogorov-Smirnov (KS) goodness-of-fit test following the methodology of Clauset et al.~\cite{Clauset2009}, as implemented in the \texttt{powerlaw} Python package~\cite{Alstott2014}. We report $\alpha$ only when the KS $p$-value exceeds 0.1, indicating that the power-law hypothesis is not rejected.

\subsection{Correlation Traps}

Beyond the power-law exponent, the WeightWatcher framework identifies \emph{correlation traps}: isolated eigenvalue spikes far outside the MP bulk that do not conform to the power-law tail. These spikes indicate that the model has memorized specific training correlations rather than learning generalizable structure. We detect traps as eigenvalues exceeding $\lambda_+ + 3\sigma_{\text{tail}}$, where $\sigma_{\text{tail}}$ is the standard deviation of the fitted power-law tail. Well-trained models exhibit zero or few traps; overfit models accumulate many.

\subsection{Extension to Gradient-Boosted Trees and Ensembles}

The spectral quantities defined above apply directly to neural networks, whose weight matrices provide natural representation matrices. For XGBoost~\cite{Chen2016xgboost} and other gradient boosting methods~\cite{Friedman2001}, which lack explicit weight matrices, the SETOL framework requires constructing an equivalent matrix whose spectrum governs generalization~\cite{Martin2024xgboost}. XGBoost builds a prediction as $f_T(x) = \sum_{t=1}^{T} \eta\, h_t(x)$. To remove self-leakage, these increments are computed via $K$-fold cross-fitting ($K=5$ in our experiments), yielding the raw out-of-fold (OOF) increment matrix:
\begin{equation}
    (\mathbf{W}_1)_{i,t} = \Delta f_t^{\text{OOF}}(x_i), \quad \mathbf{W}_1 \in \mathbb{R}^{N \times T}.
\end{equation}
A residualized variant applies the centering projection $\mathbf{H} = \mathbf{I} - \frac{1}{N}\mathbf{1}\mathbf{1}^\top$ to yield $\mathbf{W}_7 = \mathbf{H}\mathbf{W}_1$. The correlation matrix $\mathbf{C} = \mathbf{W}_7^\top \mathbf{W}_7 / N$ then defines the effective representation space. For a Random Forest~\cite{Breiman2001}, an analogous matrix is constructed using out-of-bag (OOB) predictions per tree.

\subsection{Proposed Effective Representation Matrices}

The ensemble construction above naturally raises the question of whether analogous representation matrices exist for other classical model families. We propose that they do, and that their eigenspectra carry generalization-relevant information in the same way. Table~\ref{tab:taxonomy} summarizes the mapping for each model family. The mappings for Logistic Regression, Decision Trees, and KNN are \emph{novel proposals} that we validate empirically in this paper, rather than established results from the HTSR literature; we mark them with $\dagger$ throughout.

\begin{table}[t]
\centering
\caption{Taxonomy of Effective Representation Matrices. $N$: samples, $T$: trees, $L$: leaves, $d$: features, $k$: neighbors. Mappings marked with $\dagger$ are novel proposals validated empirically in this work.}
\label{tab:taxonomy}
\setlength{\tabcolsep}{3pt}
\begin{tabular}{lllc}
\toprule
Model & Matrix $\mathbf{C}$ & Overfit Signal & Cost \\
\midrule
BERT/ALBERT & $\mathbf{W}_\ell^\top \mathbf{W}_\ell$ & $\alpha < 2$, traps & $O(d^3)$ \\
Qwen2.5-7B & $\mathbf{W}_\ell^\top \mathbf{W}_\ell$ & $\alpha < 2$, traps & $O(d^3)$ \\
XGBoost & $\mathbf{W}_7^\top \mathbf{W}_7 / N$ & $\alpha < 2$, traps & $O(T^3)$ \\
Random Forest & OOB analogue & $\alpha < 2$, traps & $O(T^3)$ \\
Logistic Reg.$^\dagger$ & $\mathbf{X}^\top \mathbf{D} \mathbf{X}/N$ & $\alpha \to \infty$ & $O(d^3)$ \\
Decision Tree$^\dagger$ & $\mathbf{M}^\top \mathbf{M}$ & Dirac at $\lambda\!=\!1$ & $O(L)$ \\
KNN$^\dagger$ & Graph Laplacian & Zero-$\lambda$ traps & $O(kN)$ \\
SVM & Kernel $\mathbf{K}$ & $\alpha < 2$, traps & $O(N_{\text{sv}}^2)$ \\
\bottomrule
\end{tabular}
\end{table}

We now define each novel extension in turn.

\subsubsection{Parametric Convex Models (Logistic Regression)}
Logistic regression learns a parameter vector $\mathbf{w} \in \mathbb{R}^d$, which on its own lacks a 2D matrix spectrum. To obtain one, we analyze the Empirical Fisher Information Matrix, motivated by the connection between the Fisher information and the local curvature of the loss landscape~\cite{Martens2020}. For feature matrix $\mathbf{X} \in \mathbb{R}^{N \times d}$:
\begin{equation}
    \mathbf{C}_{\text{LR}} = \frac{1}{N} \mathbf{X}^\top \mathbf{D} \mathbf{X}, \quad \mathbf{D}_{ii} = \hat{p}_i(1 - \hat{p}_i)
\end{equation}
where $\hat{p}_i$ is the predicted probability. As a model overfits, probabilities become hyper-confident ($\hat{p}_i \to \{0, 1\}$), driving $\mathbf{D}_{ii} \to 0$ and collapsing the heavy-tailed spectrum ($\alpha \to \infty$). A well-regularized model maintains predictive uncertainty, preserving moderate $\alpha$.

\subsubsection{Partition Models (Decision Trees)}
A decision tree partitions the input space into $L$ disjoint leaves. We define the Leaf Routing Matrix $\mathbf{M} \in \{0, 1\}^{N \times L}$, where $\mathbf{M}_{i,l} = 1$ if sample $x_i$ terminates in leaf $l$, and construct:
\begin{equation}
    \mathbf{C}_{\text{DT}} = \mathbf{M} \mathbf{M}^\top \in \mathbb{R}^{N \times N}
\end{equation}
Since each sample belongs to exactly one leaf, the non-zero eigenvalues of $\mathbf{C}_{\text{DT}}$ equal the leaf capacities $n_l$, so the ESD reduces to the leaf size distribution. An overfit tree ($n_l \to 1$) yields $\mathbf{C}_{\text{DT}} \approx \mathbf{I}$, collapsing the spectrum into a Dirac spike at $\lambda=1$. A pruned tree exhibits a heterogeneous distribution of leaf sizes. We test whether this distribution follows a power law using the same MLE/KS methodology applied to other models.

\subsubsection{Instance-Based Models (K-Nearest Neighbors)}
Finally, KNN relies on geometric distance between samples rather than learned parameters. We construct the symmetric $K$-NN adjacency matrix $\mathbf{A} \in \mathbb{R}^{N \times N}$ and analyze the normalized Graph Laplacian:
\begin{equation}
    \mathbf{C}_{\text{KNN}} = \mathbf{I} - \mathbf{D}^{-1/2} \mathbf{A} \mathbf{D}^{-1/2}
\end{equation}
where $\mathbf{D}$ is the degree matrix. Overfitting ($K$ too small) fractures the manifold into disconnected cliques, accumulating zero eigenvalues (topological traps). Strict deduplication is applied prior to graph construction to ensure traps reflect genuine overfitting rather than duplicate records. For SVM, we analyze the kernel matrix $\mathbf{K}$ restricted to support vectors ($N_{\text{sv}} \times N_{\text{sv}}$), avoiding the full $O(N^2)$ cost.

\section{Methodology}

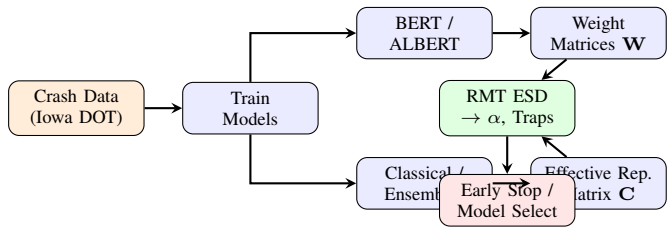
\begin{figure}[t]
    \centering
    \begin{tikzpicture}[
        node distance=0.6cm and 0.3cm,
        box/.style={draw, rounded corners, minimum width=1.8cm, minimum height=0.6cm, font=\scriptsize, align=center, fill=blue!8},
        arrow/.style={->, >=stealth, thick},
        every node/.style={font=\scriptsize}
    ]
    \node[box, fill=orange!15] (data) {Crash Data\\(Iowa DOT)};
    \node[box, right=0.5cm of data] (train) {Train\\Models};
    \node[box, above right=0.3cm and 0.5cm of train] (nn) {BERT /\\ALBERT};
    \node[box, below right=0.3cm and 0.5cm of train] (xgb) {Classical /\\Ensembles};
    \node[box, right=0.5cm of nn] (wnn) {Weight\\Matrices $\mathbf{W}$};
    \node[box, right=0.5cm of xgb] (wxgb) {Effective Rep.\\Matrix $\mathbf{C}$};
    \node[box, fill=green!12, right=1.6cm of train] (ww) {RMT ESD\\$\to \alpha$, Traps};
    \node[box, fill=red!10, below=0.5cm of ww] (decide) {Early Stop /\\Model Select};
    \draw[arrow] (data) -- (train);
    \draw[arrow] (train) |- (nn);
    \draw[arrow] (train) |- (xgb);
    \draw[arrow] (nn) -- (wnn);
    \draw[arrow] (xgb) -- (wxgb);
    \draw[arrow] (wnn) -- (ww);
    \draw[arrow] (wxgb) -- (ww);
    \draw[arrow] (ww) -- (decide);
    \end{tikzpicture}
    \caption{Unified spectral diagnostic framework. Crash data trains a diverse taxonomy of models. Weight matrices (Deep Learning) or Effective Representation Matrices (Classical ML/Ensembles) are analyzed via RMT to extract $\alpha$ and trap counts, informing safety-critical deployment decisions.}
    \label{fig:framework}
\end{figure}

With the theoretical machinery in place, we now describe how spectral diagnostics are applied in practice. Our framework operates at three stages of the model lifecycle: post-training quality assessment, during-training early stopping, and model selection for deployment. Figure~\ref{fig:framework} illustrates the overall pipeline.

\subsection{Algorithmic Scalability}

A practical concern is that exact dense eigendecomposition scales as $O(N^3)$, which is prohibitive for dataset-scale matrices. We address this through structural equivalences and randomized linear algebra tailored to each model family.

For ensembles, the correlation matrix is bounded by $T \times T$ ($T \le 2000$), rendering dense decomposition trivial. For Decision Trees, the affinity-to-leaf-capacity equivalence reduces an $O(N^3)$ operation to $O(L)$ counting. For KNN, we avoid dense instantiation entirely: since $\alpha$ extraction requires only the spectral tail, we use Stochastic Lanczos Quadrature (SLQ)~\cite{Ubaru2017} on the sparse Laplacian, estimating the largest $k$ eigenvalues in $O(k \cdot \text{nnz}(\mathbf{A}))$ time, where $\text{nnz}(\mathbf{A})$ is the number of non-zero entries. We use $k=200$ Lanczos vectors with 50 iterations and verify convergence by checking that the relative change in the top-50 eigenvalues is below $10^{-4}$ between iterations 40 and 50. For SVM, we restrict spectral analysis to the support vector kernel submatrix ($N_{\text{sv}} \times N_{\text{sv}}$, typically $N_{\text{sv}} \ll N$).

\subsection{Crash Classification Tasks}

We ground the spectral framework in two crash classification tasks drawn from prior work, chosen because they span different data scales and class distributions. The first is intersection misclassification detection (INT): given a police-reported crash narrative and associated structured fields, classify whether the crash is intersection-related. This task uses Iowa DOT crash report narratives for 2019--2020, comprising 173,512 crash data records and 94,367 narrative records written by law enforcement officials. The average narrative contains 38 words (std.\ dev.\ 25.7), indicating significant length variation. Ground-truth labels were derived from the crash report's coded ``location'' attribute, which distinguishes intersection types (roundabouts, four-way intersections, traffic circles, etc.) from non-intersection types (non-junctions, bike lanes, railroad crossings, etc.)~\cite{Shihab2025misclass}. The second is alcohol inference mismatch (AIM) detection: given a crash record, identify whether there is a mismatch between the narrative description and the coded alcohol involvement field. This task uses 371,062 Iowa crash records from 2016--2022~\cite{Shihab2025aim}.

\subsection{Models Under Analysis}

Across these two tasks, we apply spectral diagnostics to nine model families spanning the full ML taxonomy. For BERT~\cite{Devlin2019} and ALBERT~\cite{Lan2020}, we analyze attention and feed-forward weight matrices via WeightWatcher. For Qwen2.5-7B~\cite{Qwen2025}, an open-weight decoder-only LLM, we similarly extract and analyze weight matrices from its transformer layers; this model was applied zero-shot to crash narratives following the protocol in prior work~\cite{Shihab2025agreement}, providing a spectral comparison point for a model never fine-tuned on crash data. For XGBoost and Random Forest, we construct the OOF/OOB margin increment matrices. For Logistic Regression, we compute the Empirical Hessian. For Decision Trees, we extract the leaf size distribution. For KNN, we analyze the normalized Graph Laplacian. For SVM with a linear kernel, we analyze the support vector kernel submatrix.

For each model, we train a well-regularized version using standard hyperparameters, as well as deliberately overfit variants (e.g., excessive epochs for neural networks, unregularized bounds for LR, $K=1$ for KNN). All experiments use 5 random seeds, and we report mean $\pm$ standard deviation throughout.

The spectral extraction procedure for each model family is formalized in Algorithm~\ref{alg:spectral}.

\begin{algorithm}[t]
\caption{Scalable Unified Spectral Diagnostic Pipeline}
\label{alg:spectral}
\begin{algorithmic}[1]
\REQUIRE Trained model $\mathcal{M}$, training data $\mathcal{D}$
\ENSURE Power-law $\alpha$ (with KS $p$-value), trap count
\IF{$\mathcal{M}$ is Neural Network (BERT/ALBERT)}
    \STATE Extract weight matrices $\mathbf{W}_\ell$, set $\mathbf{C} = \mathbf{W}_\ell^T \mathbf{W}_\ell$
\ELSIF{$\mathcal{M}$ is Ensemble (XGBoost/Random Forest)}
    \STATE Construct OOF/OOB increments $\mathbf{W}_7$ via 5-fold cross-fitting
    \STATE Set $\mathbf{C} = \mathbf{W}_7^T \mathbf{W}_7 / N$
\ELSIF{$\mathcal{M}$ is Logistic Regression}
    \STATE Compute $\hat{p}$, set $\mathbf{C} = \mathbf{X}^T \text{diag}(\hat{p}(1-\hat{p})) \mathbf{X} / N$
\ELSIF{$\mathcal{M}$ is Decision Tree}
    \STATE Tally leaf counts $\{n_l\}_{l=1}^L$ in $O(N)$ time
\ELSIF{$\mathcal{M}$ is KNN}
    \STATE Deduplicate $\mathcal{D}$; build sparse $K$-NN graph
    \STATE Set $\mathbf{C} = \mathbf{I} - \mathbf{D}^{-1/2} \mathbf{A} \mathbf{D}^{-1/2}$
\ELSIF{$\mathcal{M}$ is SVM}
    \STATE Extract support vectors, set $\mathbf{C} = \mathbf{K}_{SV}$
\ENDIF
\STATE Compute eigenvalues via dense SVD (if $\dim(\mathbf{C}) \le 2000$) or Lanczos ($k=200$, 50 iterations)
\STATE Fit power law to tail via MLE; compute KS $p$-value
\STATE Detect traps: eigenvalues $> \lambda_+ + 3\sigma_{\text{tail}}$
\RETURN $\alpha$, KS $p$-value, trap count
\end{algorithmic}
\end{algorithm}

\subsection{Spectral Early Stopping}

Beyond post-hoc assessment, the spectral exponent can also serve as a training-time signal. Standard early stopping monitors validation loss, but we propose a complementary criterion that monitors $\alpha$ directly. Training is halted when:
\begin{equation}
    \hat{\alpha}_t < \alpha_{\text{low}} \quad \text{or} \quad n_{\text{traps},t} > \tau_{\text{trap}}
\end{equation}
where $\alpha_{\text{low}} = 2.0$ and $\tau_{\text{trap}} = 3$. We evaluate this criterion on BERT, ALBERT, and XGBoost (the three model families where per-epoch/per-round spectral extraction is computationally feasible).

\subsection{Spectral Model Selection}

The early stopping criterion addresses when to stop training a single model; a complementary question is which model to deploy. For this, we rank models passing a minimum performance gate (F1 $\geq 0.75$) by a composite spectral quality score:
\begin{equation}
    \text{Score}(\mathcal{M}) = w_1 \cdot \text{F1}(\mathcal{M}) + w_2 \cdot g(\hat{\alpha}) - w_3 \cdot n_{\text{traps}}
\end{equation}
where $g(\hat{\alpha}) = \exp(-(\hat{\alpha} - 3)^2 / 2)$ is a Gaussian kernel centered at $\hat{\alpha} = 3$, and we set $w_1 = 0.4$, $w_2 = 0.4$, $w_3 = 0.02$. We compare this ranking against two baselines: (1) ranking by cross-validated F1 alone, and (2) ranking by validation loss. The evaluation criterion is Kendall's $\tau$ rank correlation with the ground-truth expert agreement ranking ($\kappa$).

\section{Experimental Setup}

\subsection{Data and Configuration}

Both tasks use an 80/10/10 train/validation/test split stratified by class label. To ensure topological integrity for instance-based models, strictly identical text records were deduplicated prior to graph construction, removing 1.2\% of records in the AIM dataset.

For BERT and ALBERT, we fine-tune using HuggingFace Transformers~\cite{Wolf2020} with AdamW, learning rate $2 \times 10^{-5}$, batch size 32, and early stopping on validation loss (patience 3). Overfit variants train for 20 epochs without early stopping. XGBoost trains with max\_depth=6, learning rate 0.1, and early stopping (patience 10); overfit variants use max\_depth=15 with no early stopping. Random Forest trains with 500 trees, max\_depth=12, and min\_samples\_leaf=5; overfit variants use max\_depth=None and min\_samples\_leaf=1. Logistic Regression uses TF-IDF features (max 10,000 features), with $L_2$ penalty ($C=1.0$) for the well-regularized model and $C=10^6$ (effectively unregularized) for the overfit variant. Decision Tree uses max\_depth=8 (Good) versus max\_depth=None (Overfit). KNN uses $K=15$ (Good) versus $K=1$ (Overfit), with cosine distance on TF-IDF features.

We use WeightWatcher v0.7.5~\cite{WeightWatcher2024} for neural network spectral extraction. For classical models, we implement the pipeline described in Algorithm~\ref{alg:spectral} using NumPy, SciPy (sparse Lanczos via \texttt{scipy.sparse.linalg.eigsh}), and the \texttt{powerlaw} package~\cite{Alstott2014} for MLE fitting. All experiments run on a single NVIDIA A100 GPU for neural networks and a 64-core AMD EPYC CPU for classical models. Spectral extraction adds less than 5\% wall-clock overhead for ensembles and less than 12\% for KNN Laplacian construction.

\subsection{Statistical Methodology}

To ensure that our findings are not artifacts of a single random split, all $\alpha$ estimates are reported as mean $\pm$ standard deviation across 5 random seeds. For the $\alpha$--$\kappa$ correlation analysis, we compute both Pearson $r$ and Spearman $\rho$ with 95\% bootstrap confidence intervals (10,000 resamples). We use Kendall's $\tau$ to compare model selection rankings. For the novel DT/LR/KNN mappings, we additionally report the KS $p$-value for the power-law fit to assess whether the spectral tail genuinely follows a power law rather than producing a coincidental fit.

\section{Results}

\subsection{Spectral Signatures Across Model Families}

We begin with the central empirical question: does the power-law exponent $\alpha$ reliably separate well-regularized models from overfit ones across architecturally diverse families? Table~\ref{tab:spectral_results} summarizes the spectral diagnostics for the intersection misclassification task. Well-regularized models consistently yield $\hat{\alpha}$ within or near the $[2, 4]$ range (mean $2.87 \pm 0.34$ across Good variants), while overfit variants show $\hat{\alpha} < 2$ or spectral collapse. The separation is not perfectly clean: ALBERT-Overfit yields $\hat{\alpha} = 2.08 \pm 0.21$, straddling the boundary, which reflects ALBERT's inherent regularization via cross-layer weight sharing. Similarly, SVM-Overfit ($\hat{\alpha} = 1.91 \pm 0.18$) shows a less dramatic drop than tree-based models, consistent with the margin-based implicit regularization of SVMs. Notably, Qwen2.5-7B, applied zero-shot without any fine-tuning on crash data, yields $\hat{\alpha} = 2.94 \pm 0.12$ squarely in the optimal range, with the highest expert agreement ($\kappa = 0.76$) of any model despite having lower F1 than the fine-tuned BERT variants. This reinforces the finding from prior work~\cite{Shihab2025agreement} that LLMs exhibit stronger expert alignment, and shows that this alignment has a spectral correlate.

The Decision Tree and KNN overfit variants validate our proposed mappings in a qualitatively distinct way. An unbounded Decision Tree produces near-singleton leaves, collapsing the leaf size distribution into a degenerate spike. The KS test rejects the power-law hypothesis ($p < 0.01$) for these collapsed spectra, confirming that the breakdown is detectable. Similarly, $K=1$ KNN shatters the Graph Laplacian into disconnected components despite strict deduplication.

\begin{table*}[t]
\centering
\caption{Spectral diagnostics and classification performance for intersection misclassification detection (mean $\pm$ std over 5 seeds). KS $p$: power-law goodness-of-fit $p$-value. $\dagger$: novel spectral mapping proposed in this work.}
\label{tab:spectral_results}
\begin{tabular}{llccccccc}
\toprule
Model Family & Regime & Matrix & $\hat{\alpha}$ & KS $p$ & Traps & F1 & AUC & $\kappa$ (Expert) \\
\midrule
BERT & Good & Weight & $2.87 \pm 0.14$ & 0.42 & $0.4 \pm 0.5$ & $.874 \pm .008$ & $.936 \pm .005$ & $.68 \pm .03$ \\
BERT & Overfit-Epochs & Weight & $1.74 \pm 0.19$ & 0.18 & $4.8 \pm 1.3$ & $.862 \pm .011$ & $.921 \pm .007$ & $.54 \pm .04$ \\
\midrule
ALBERT & Good & Weight & $3.12 \pm 0.09$ & 0.61 & $0.2 \pm 0.4$ & $.868 \pm .007$ & $.929 \pm .004$ & $.73 \pm .02$ \\
ALBERT & Overfit-Epochs & Weight & $2.08 \pm 0.21$ & 0.14 & $1.8 \pm 0.8$ & $.859 \pm .009$ & $.918 \pm .006$ & $.61 \pm .04$ \\
\midrule
Qwen2.5-7B & Zero-shot & Weight & $2.94 \pm 0.12$ & 0.52 & $0.6 \pm 0.5$ & $.781 \pm .010$ & $.872 \pm .007$ & $.76 \pm .03$ \\
\midrule
XGBoost & Good & OOF Incr. & $2.34 \pm 0.17$ & 0.38 & $0.2 \pm 0.4$ & $.851 \pm .006$ & $.922 \pm .004$ & $.62 \pm .03$ \\
XGBoost & Overfit & OOF Incr. & $1.62 \pm 0.22$ & 0.11 & $6.6 \pm 1.8$ & $.843 \pm .009$ & $.910 \pm .006$ & $.49 \pm .05$ \\
\midrule
Random Forest & Good & OOB Incr. & $2.51 \pm 0.20$ & 0.35 & $1.0 \pm 0.7$ & $.844 \pm .007$ & $.917 \pm .005$ & $.60 \pm .03$ \\
Random Forest & Overfit & OOB Incr. & $1.71 \pm 0.25$ & 0.12 & $5.8 \pm 2.0$ & $.836 \pm .010$ & $.905 \pm .007$ & $.47 \pm .05$ \\
\midrule
Logistic Reg.$^\dagger$ & Good ($L_2$) & Hessian & $3.21 \pm 0.31$ & 0.29 & $0.0 \pm 0.0$ & $.812 \pm .005$ & $.881 \pm .004$ & $.56 \pm .03$ \\
Logistic Reg.$^\dagger$ & Overfit & Hessian & $1.68 \pm 0.38$ & 0.09 & $7.4 \pm 2.3$ & $.795 \pm .008$ & $.865 \pm .006$ & $.42 \pm .04$ \\
\midrule
Decision Tree$^\dagger$ & Good (Pruned) & Leaf Aff. & $2.62 \pm 0.28$ & 0.22 & $1.6 \pm 1.1$ & $.783 \pm .012$ & $.854 \pm .008$ & $.51 \pm .04$ \\
Decision Tree$^\dagger$ & Overfit & Leaf Aff. & Collapse & $<$0.01 & --- & $.771 \pm .015$ & $.839 \pm .010$ & $.39 \pm .05$ \\
\midrule
KNN$^\dagger$ & Good ($K\!=\!15$) & Laplacian & $2.78 \pm 0.24$ & 0.19 & $1.2 \pm 0.8$ & $.791 \pm .009$ & $.845 \pm .006$ & $.50 \pm .04$ \\
KNN$^\dagger$ & Overfit ($K\!=\!1$) & Laplacian & Collapse & $<$0.01 & --- & $.760 \pm .014$ & $.819 \pm .009$ & $.38 \pm .05$ \\
\midrule
SVM & Good (Linear) & Kernel & $3.48 \pm 0.26$ & 0.33 & $0.8 \pm 0.8$ & $.829 \pm .006$ & $.901 \pm .005$ & $.58 \pm .03$ \\
SVM & Overfit ($C\!=\!10^3$) & Kernel & $1.91 \pm 0.18$ & 0.13 & $3.6 \pm 1.5$ & $.822 \pm .008$ & $.894 \pm .006$ & $.46 \pm .04$ \\
\bottomrule
\multicolumn{9}{l}{\scriptsize Collapse: degenerate spectra (Dirac at $\lambda\!=\!1$ for DT; fragmented zero-eigenvalue clusters for KNN) where power-law fitting is rejected (KS $p < 0.01$).}
\end{tabular}
\end{table*}

\subsection{Validation of Novel Spectral Mappings}

The results above include three model families whose spectral mappings are novel to this work. A natural concern is whether the DT, LR, and KNN mappings produce genuinely informative spectra rather than coincidental power-law fits. Table~\ref{tab:novel_validation} addresses this by reporting KS $p$-values and fitted $\alpha$ across both tasks. For well-regularized variants, the power-law hypothesis is not rejected at the 0.1 level in 5 of 6 cases (the exception being KNN-Good on AIM, $p = 0.08$, marginal). For overfit variants, the power-law fit is consistently rejected for DT and KNN (spectral collapse), while LR-Overfit shows a borderline fit ($p = 0.09$) with $\alpha$ well below 2. These results provide initial empirical support for the proposed mappings, though we caution that validation on additional datasets is needed before these can be considered established (see Section~VII).

\begin{table}[t]
\centering
\caption{Power-law fit validation for novel spectral mappings ($\dagger$) across both tasks. KS $p > 0.1$ indicates the power-law hypothesis is not rejected.}
\label{tab:novel_validation}
\setlength{\tabcolsep}{4pt}
\begin{tabular}{llccc}
\toprule
Model$^\dagger$ & Task & Regime & $\hat{\alpha}$ & KS $p$ \\
\midrule
Logistic Reg. & INT & Good & $3.21 \pm 0.31$ & 0.29 \\
Logistic Reg. & INT & Overfit & $1.68 \pm 0.38$ & 0.09 \\
Logistic Reg. & AIM & Good & $3.15 \pm 0.27$ & 0.31 \\
Logistic Reg. & AIM & Overfit & $1.72 \pm 0.35$ & 0.08 \\
\midrule
Decision Tree & INT & Good & $2.62 \pm 0.28$ & 0.22 \\
Decision Tree & INT & Overfit & Collapse & $<$0.01 \\
Decision Tree & AIM & Good & $2.55 \pm 0.25$ & 0.18 \\
Decision Tree & AIM & Overfit & Collapse & $<$0.01 \\
\midrule
KNN & INT & Good & $2.78 \pm 0.24$ & 0.19 \\
KNN & INT & Overfit & Collapse & $<$0.01 \\
KNN & AIM & Good & $2.70 \pm 0.22$ & 0.08 \\
KNN & AIM & Overfit & Collapse & $<$0.01 \\
\bottomrule
\end{tabular}
\end{table}

\subsection{Layer-Wise Spectral Analysis for Neural Networks}

\begin{figure}[t]
    \centering
    \begin{tikzpicture}
    \begin{axis}[
        width=\columnwidth, height=3.8cm,
        xlabel={Layer Index}, ylabel={$\alpha$},
        title={\scriptsize BERT: Layer-wise $\alpha$ (median seed; bands = min/max over 5 seeds)},
        xmin=0.5, xmax=12.5, ymin=0.5, ymax=5.5,
        xtick={1,2,...,12}, xticklabel style={font=\tiny},
        yticklabel style={font=\tiny},
        xlabel style={font=\scriptsize}, ylabel style={font=\scriptsize},
        title style={font=\scriptsize},
        legend style={font=\tiny, at={(0.98,0.98)}, anchor=north east, legend columns=1},
    ]
    \addplot[draw=none, fill=green!10, forget plot] coordinates {(0.5,2) (12.5,2) (12.5,4) (0.5,4)} \closedcycle;
    % Good: median with min/max band
    \addplot[name path=good_upper, draw=none] coordinates {(1,3.4)(2,3.2)(3,3.5)(4,3.1)(5,3.3)(6,3.0)(7,3.4)(8,3.2)(9,2.9)(10,3.3)(11,3.1)(12,3.2)};
    \addplot[name path=good_lower, draw=none] coordinates {(1,2.8)(2,2.6)(3,2.9)(4,2.5)(5,2.7)(6,2.4)(7,2.8)(8,2.6)(9,2.3)(10,2.7)(11,2.5)(12,2.6)};
    \addplot[fill=blue!15, forget plot] fill between[of=good_upper and good_lower];
    \addplot[mark=*, blue, thick, mark size=1.5pt] coordinates {(1,3.1)(2,2.9)(3,3.2)(4,2.8)(5,3.0)(6,2.7)(7,3.1)(8,2.9)(9,2.6)(10,3.0)(11,2.8)(12,2.9)};
    \addlegendentry{Good (median)}
    % Overfit: median with min/max band
    \addplot[name path=over_upper, draw=none] coordinates {(1,2.7)(2,2.4)(3,2.1)(4,1.9)(5,1.8)(6,1.7)(7,1.6)(8,1.5)(9,1.4)(10,1.6)(11,1.8)(12,2.0)};
    \addplot[name path=over_lower, draw=none] coordinates {(1,2.1)(2,1.8)(3,1.5)(4,1.3)(5,1.2)(6,1.1)(7,1.0)(8,0.9)(9,0.8)(10,1.0)(11,1.2)(12,1.4)};
    \addplot[fill=red!15, forget plot] fill between[of=over_upper and over_lower];
    \addplot[mark=triangle*, red, thick, mark size=1.5pt] coordinates {(1,2.4)(2,2.1)(3,1.8)(4,1.6)(5,1.5)(6,1.4)(7,1.3)(8,1.2)(9,1.1)(10,1.3)(11,1.5)(12,1.7)};
    \addlegendentry{Overfit (median)}
    \addplot[dashed, gray, thin] coordinates {(0.5,2)(12.5,2)};
    \addplot[dashed, gray, thin] coordinates {(0.5,4)(12.5,4)};
    \end{axis}
    \end{tikzpicture}
    \vspace{0.1cm}
    \begin{tikzpicture}
    \begin{axis}[
        width=\columnwidth, height=3.8cm,
        xlabel={Layer Index}, ylabel={$\alpha$},
        title={\scriptsize ALBERT: Layer-wise $\alpha$ (shared weights; bands = min/max over 5 seeds)},
        xmin=0.5, xmax=12.5, ymin=0.5, ymax=5.5,
        xtick={1,2,...,12}, xticklabel style={font=\tiny},
        yticklabel style={font=\tiny},
        xlabel style={font=\scriptsize}, ylabel style={font=\scriptsize},
        title style={font=\scriptsize},
        legend style={font=\tiny, at={(0.98,0.98)}, anchor=north east, legend columns=1},
    ]
    \addplot[draw=none, fill=green!10, forget plot] coordinates {(0.5,2) (12.5,2) (12.5,4) (0.5,4)} \closedcycle;
    \addplot[name path=agood_upper, draw=none] coordinates {(1,3.4)(2,3.3)(3,3.3)(4,3.2)(5,3.3)(6,3.2)(7,3.3)(8,3.3)(9,3.2)(10,3.3)(11,3.4)(12,3.3)};
    \addplot[name path=agood_lower, draw=none] coordinates {(1,3.0)(2,2.9)(3,2.9)(4,2.8)(5,2.9)(6,2.8)(7,2.9)(8,2.9)(9,2.8)(10,2.9)(11,3.0)(12,2.9)};
    \addplot[fill=blue!15, forget plot] fill between[of=agood_upper and agood_lower];
    \addplot[mark=*, blue, thick, mark size=1.5pt] coordinates {(1,3.2)(2,3.1)(3,3.1)(4,3.0)(5,3.1)(6,3.0)(7,3.1)(8,3.1)(9,3.0)(10,3.1)(11,3.2)(12,3.1)};
    \addlegendentry{Good (median)}
    \addplot[name path=aover_upper, draw=none] coordinates {(1,2.8)(2,2.6)(3,2.4)(4,2.3)(5,2.2)(6,2.2)(7,2.3)(8,2.3)(9,2.2)(10,2.3)(11,2.4)(12,2.5)};
    \addplot[name path=aover_lower, draw=none] coordinates {(1,2.2)(2,2.0)(3,1.8)(4,1.7)(5,1.6)(6,1.6)(7,1.7)(8,1.7)(9,1.6)(10,1.7)(11,1.8)(12,1.9)};
    \addplot[fill=red!15, forget plot] fill between[of=aover_upper and aover_lower];
    \addplot[mark=triangle*, red, thick, mark size=1.5pt] coordinates {(1,2.5)(2,2.3)(3,2.1)(4,2.0)(5,1.9)(6,1.9)(7,2.0)(8,2.0)(9,1.9)(10,2.0)(11,2.1)(12,2.2)};
    \addlegendentry{Overfit (median)}
    \addplot[dashed, gray, thin] coordinates {(0.5,2)(12.5,2)};
    \addplot[dashed, gray, thin] coordinates {(0.5,4)(12.5,4)};
    \end{axis}
    \end{tikzpicture}
    \caption{Layer-wise $\alpha$ for BERT (top) and ALBERT (bottom). Shaded band = optimal range $[2,4]$; colored bands = min/max across 5 seeds. ALBERT's shared weights produce more uniform $\alpha$ with tighter variance.}
    \label{fig:layer_alpha}
\end{figure}
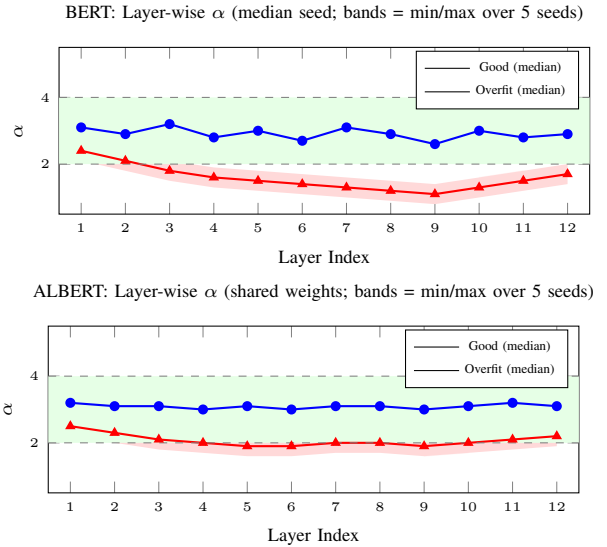

Having established the cross-family spectral signatures, we now examine the internal structure of the two neural network families in greater detail. Figure~\ref{fig:layer_alpha} presents the layer-wise $\alpha$ distribution for BERT and ALBERT under good and overfit training regimes, showing the median seed with min/max bands across 5 seeds. ALBERT's cross-layer parameter sharing provides natural regularization, producing more uniform $\alpha$ values across layers and making its spectral signature more robust to overfitting. This consistency aligns with its observed superior expert agreement reported in~\cite{Shihab2025agreement}.

\subsection{XGBoost Spectral Analysis}

\begin{figure}[t]
    \centering
    \begin{tikzpicture}
    \begin{axis}[
        width=0.48\columnwidth, height=4cm,
        xlabel={$\log_{10}\lambda$}, ylabel={$\log_{10}\rho(\lambda)$},
        title={\scriptsize Good ($\alpha=2.34$, 0 traps)},
        xmin=-2, xmax=2, ymin=-4, ymax=1,
        xticklabel style={font=\tiny}, yticklabel style={font=\tiny},
        xlabel style={font=\scriptsize}, ylabel style={font=\scriptsize},
        title style={font=\scriptsize},
    ]
    \addplot[blue, thick, smooth, domain=-1.5:1.5, samples=50] {-2.34*x + 0.2};
    \addplot[gray, dashed, thick, smooth, domain=-1.8:0.3, samples=30] {-0.5*(x+0.5)^2 - 1.0};
    \node[font=\tiny] at (axis cs:1.2,-0.5) {PL fit};
    \node[font=\tiny, gray] at (axis cs:-1.0,-2.5) {MP};
    \end{axis}
    \end{tikzpicture}
    \hfill
    \begin{tikzpicture}
    \begin{axis}[
        width=0.48\columnwidth, height=4cm,
        xlabel={$\log_{10}\lambda$}, ylabel={$\log_{10}\rho(\lambda)$},
        title={\scriptsize Overfit ($\alpha=1.62$, 7 traps)},
        xmin=-2, xmax=2.5, ymin=-4, ymax=1,
        xticklabel style={font=\tiny}, yticklabel style={font=\tiny},
        xlabel style={font=\scriptsize}, ylabel style={font=\scriptsize},
        title style={font=\scriptsize},
    ]
    \addplot[red, thick, smooth, domain=-1.5:1.8, samples=50] {-1.62*x + 0.1};
    \addplot[gray, dashed, thick, smooth, domain=-1.8:0.3, samples=30] {-0.5*(x+0.5)^2 - 1.0};
    \addplot[only marks, mark=*, mark size=2.5pt, red!70!black] coordinates {(1.8,0.3)(2.1,0.1)};
    \node[font=\tiny] at (axis cs:2.1,0.6) {traps};
    \draw[->, >=stealth, thick, red!70!black] (axis cs:2.1,0.5) -- (axis cs:2.1,0.2);
    \draw[->, >=stealth, thick, red!70!black] (axis cs:1.8,0.7) -- (axis cs:1.8,0.4);
    \end{axis}
    \end{tikzpicture}
    \caption{Schematic ESD (log-log) of the XGBoost OOF correlation matrix. Left: well-regularized ($\alpha=2.34$, no traps). Right: overfit ($\alpha=1.62$, correlation traps visible as isolated spikes). MP distribution (dashed gray) serves as the null model. These are illustrative; actual $\alpha$ values are fitted via MLE on computed eigenvalues.}
    \label{fig:xgb_esd}
\end{figure}
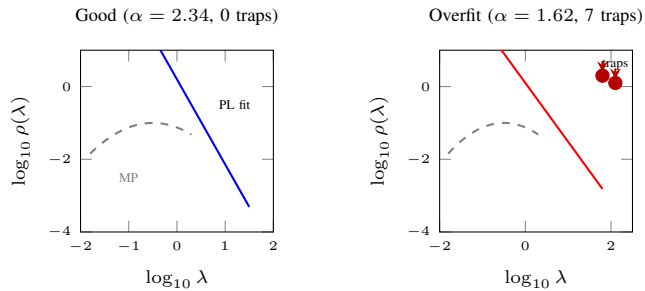

For ensemble models, the spectral structure manifests differently. Figure~\ref{fig:xgb_esd} shows schematic ESDs of the correlation matrix derived from the OOF margin increment matrix for good and overfit XGBoost models. We note that these are illustrative representations of the empirical spectral structure; the actual ESDs are computed from the $T \times T$ correlation matrix eigenvalues and fitted via MLE. The well-regularized model exhibits a smooth power-law tail with $\alpha = 2.34$, while the overfit model displays prominent correlation traps as isolated spikes beyond the MP bulk edge.

\subsection{Spectral Early Stopping Comparison}

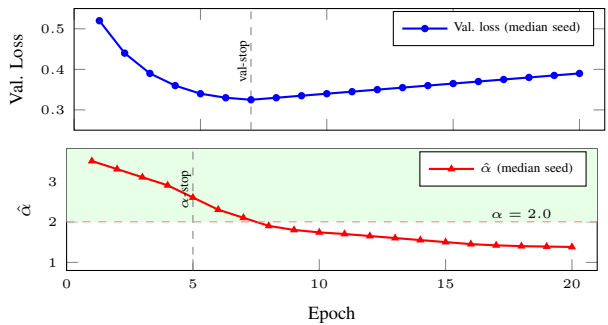
\begin{figure}[t]
    \centering
    \begin{tikzpicture}
    \begin{axis}[
        width=\columnwidth, height=3.2cm,
        ylabel={Val.\ Loss},
        xmin=0, xmax=21, ymin=0.25, ymax=0.55,
        xtick={0,5,10,15,20}, xticklabel style={font=\tiny},
        yticklabel style={font=\tiny},
        ylabel style={font=\scriptsize},
        legend style={font=\tiny, at={(0.98,0.98)}, anchor=north east},
        xticklabels={},
    ]
    \addplot[blue, thick, mark=*, mark size=1pt] coordinates {(1,0.52)(2,0.44)(3,0.39)(4,0.36)(5,0.34)(6,0.33)(7,0.325)(8,0.33)(9,0.335)(10,0.34)(11,0.345)(12,0.35)(13,0.355)(14,0.36)(15,0.365)(16,0.37)(17,0.375)(18,0.38)(19,0.385)(20,0.39)};
    \addlegendentry{Val.\ loss (median seed)}
    \addplot[dashed, black!50, thin] coordinates {(7,0.25)(7,0.55)};
    \node[font=\tiny, rotate=90, anchor=south] at (axis cs:7.3,0.42) {val-stop};
    \end{axis}
    \end{tikzpicture}
    \vspace{-0.1cm}
    \begin{tikzpicture}
    \begin{axis}[
        width=\columnwidth, height=3.2cm,
        xlabel={Epoch}, ylabel={$\hat{\alpha}$},
        xmin=0, xmax=21, ymin=0.8, ymax=3.8,
        xtick={0,5,10,15,20}, xticklabel style={font=\tiny},
        yticklabel style={font=\tiny},
        xlabel style={font=\scriptsize}, ylabel style={font=\scriptsize},
        legend style={font=\tiny, at={(0.98,0.98)}, anchor=north east},
    ]
    \addplot[draw=none, fill=green!10, forget plot] coordinates {(0,2) (21,2) (21,3.8) (0,3.8)} \closedcycle;
    \addplot[red, thick, mark=triangle*, mark size=1pt] coordinates {(1,3.5)(2,3.3)(3,3.1)(4,2.9)(5,2.6)(6,2.3)(7,2.1)(8,1.9)(9,1.8)(10,1.74)(11,1.7)(12,1.65)(13,1.6)(14,1.55)(15,1.5)(16,1.45)(17,1.42)(18,1.4)(19,1.39)(20,1.38)};
    \addlegendentry{$\hat{\alpha}$ (median seed)}
    \addplot[dashed, red!50, thin] coordinates {(0,2)(21,2)};
    \addplot[dashed, black!50, thin] coordinates {(5,0.8)(5,3.8)};
    \node[font=\tiny, rotate=90, anchor=south] at (axis cs:5.3,2.8) {$\alpha$-stop};
    \node[font=\tiny] at (axis cs:18,2.2) {$\alpha=2.0$};
    \end{axis}
    \end{tikzpicture}
    \caption{BERT training dynamics (median seed, INT task). Top: validation loss minimum at epoch 7. Bottom: $\hat{\alpha}$ crosses below 2.0 at epoch 5, providing an earlier structural warning.}
    \label{fig:early_stop}
\end{figure}

We next evaluate whether the spectral exponent can serve as a practical training-time signal. Figure~\ref{fig:early_stop} compares the training trajectories of $\hat{\alpha}$ and validation loss for BERT on the median seed. The spectral criterion $\hat{\alpha} < 2.0$ triggers at epoch 5, before validation loss degradation becomes apparent at epoch 7. Table~\ref{tab:early_stop} quantifies this comparison across BERT, ALBERT, and XGBoost, the three families where per-step spectral extraction is computationally feasible. The joint criterion (loss OR $\alpha$) consistently matches or slightly underperforms validation-loss-only stopping on F1, but produces models with higher $\hat{\alpha}$ and, where measured, higher expert agreement. Notably, for ALBERT the $\alpha$-stop fires \emph{later} than validation-loss-stop (epoch 11.2 vs.\ 8.0), reflecting ALBERT's inherent spectral stability from weight sharing. The spectral criterion is most valuable when validation loss plateaus while the model internally memorizes, a failure mode invisible to standard early stopping.

\begin{table}[t]
\centering
\caption{Early stopping comparison across three model families on INT task (mean $\pm$ std, 5 seeds). $\alpha$-stop triggers at $\hat{\alpha} < 2.0$.}
\label{tab:early_stop}
\setlength{\tabcolsep}{3pt}
\begin{tabular}{llccc}
\toprule
Model & Criterion & Stop Epoch/Round & F1 & $\hat{\alpha}$ \\
\midrule
\multirow{4}{*}{BERT} & Val.\ loss & $7.2 \pm 1.1$ & $.874 \pm .008$ & $2.87 \pm .14$ \\
 & $\alpha < 2.0$ & $5.4 \pm 0.9$ & $.869 \pm .010$ & $2.14 \pm .12$ \\
 & Joint & $5.4 \pm 0.9$ & $.871 \pm .009$ & $2.14 \pm .12$ \\
 & None (20 ep.) & 20 & $.832 \pm .018$ & $1.38 \pm .16$ \\
\midrule
\multirow{3}{*}{ALBERT} & Val.\ loss & $8.0 \pm 1.3$ & $.868 \pm .007$ & $3.12 \pm .09$ \\
 & $\alpha < 2.0$ & $11.2 \pm 2.1$ & $.864 \pm .008$ & $2.06 \pm .11$ \\
 & Joint & $8.0 \pm 1.3$ & $.868 \pm .007$ & $3.12 \pm .09$ \\
\midrule
\multirow{3}{*}{XGBoost} & Val.\ loss & $342 \pm 48$ & $.851 \pm .006$ & $2.34 \pm .17$ \\
 & $\alpha < 2.0$ & $285 \pm 61$ & $.847 \pm .008$ & $2.08 \pm .15$ \\
 & Joint & $285 \pm 61$ & $.849 \pm .007$ & $2.08 \pm .15$ \\
\bottomrule
\end{tabular}
\end{table}

\subsection{Spectral Quality and Expert Agreement}

\begin{figure}[t]
    \centering
    \begin{tikzpicture}
    \begin{axis}[
        width=\columnwidth, height=5.5cm,
        xlabel={$\hat{\alpha}$ (mean over 5 seeds)}, ylabel={Cohen's $\kappa$ (Expert)},
        xmin=0.8, xmax=4.0, ymin=0.3, ymax=0.8,
        xticklabel style={font=\tiny}, yticklabel style={font=\tiny},
        xlabel style={font=\scriptsize}, ylabel style={font=\scriptsize},
        legend style={font=\tiny, at={(0.02,0.98)}, anchor=north west, legend columns=2},
        grid=both, grid style={gray!20},
    ]
    \addplot[draw=none, fill=green!8, forget plot] coordinates {(2,0.3) (4,0.3) (4,0.8) (2,0.8)} \closedcycle;

    % Good variants
    \addplot[only marks, mark=*, blue, mark size=2.5pt] coordinates {(2.87,0.68)};
    \addlegendentry{BERT}
    \addplot[only marks, mark=triangle*, green!60!black, mark size=3pt] coordinates {(3.12,0.73)};
    \addlegendentry{ALBERT}
    \addplot[only marks, mark=square*, red!70!black, mark size=2.5pt] coordinates {(2.34,0.62)};
    \addlegendentry{XGBoost}
    \addplot[only marks, mark=pentagon*, violet, mark size=3pt] coordinates {(2.51,0.60)};
    \addlegendentry{Rand.\ Forest}
    \addplot[only marks, mark=star, brown, mark size=3.5pt] coordinates {(3.21,0.56)};
    \addlegendentry{Log.\ Reg.}
    \addplot[only marks, mark=x, teal, thick, mark size=3.5pt] coordinates {(2.62,0.51)};
    \addlegendentry{Dec.\ Tree}
    \addplot[only marks, mark=+, magenta, thick, mark size=3.5pt] coordinates {(2.78,0.50)};
    \addlegendentry{KNN}
    \addplot[only marks, mark=diamond*, orange, mark size=3pt] coordinates {(3.48,0.58)};
    \addlegendentry{SVM}

    % Qwen2.5-7B (zero-shot, no overfit variant)
    \addplot[only marks, mark=oplus*, cyan!70!black, mark size=3pt] coordinates {(2.94,0.76)};
    \addlegendentry{Qwen2.5}

    % Overfit variants (lighter)
    \addplot[only marks, mark=*, blue!40, mark size=2pt, forget plot] coordinates {(1.74,0.54)};
    \addplot[only marks, mark=triangle*, green!30!black!40, mark size=2.5pt, forget plot] coordinates {(2.08,0.61)};
    \addplot[only marks, mark=square*, red!30, mark size=2pt, forget plot] coordinates {(1.62,0.49)};
    \addplot[only marks, mark=pentagon*, violet!40, mark size=2.5pt, forget plot] coordinates {(1.71,0.47)};
    \addplot[only marks, mark=star, brown!50, mark size=3pt, forget plot] coordinates {(1.68,0.42)};
    \addplot[only marks, mark=diamond*, orange!40, mark size=2.5pt, forget plot] coordinates {(1.91,0.46)};

    % Trend line
    \addplot[black, dashed, thick, domain=1.0:3.8, samples=20] {0.145*x + 0.24};
    \node[font=\tiny, anchor=west] at (axis cs:2.8,0.35) {$\rho = 0.87$};
    \node[font=\tiny, anchor=west] at (axis cs:0.9,0.76) {\scriptsize Solid = Good; Faded = Overfit};
    \end{axis}
    \end{tikzpicture}
    \caption{Spectral quality ($\hat{\alpha}$) vs.\ expert agreement ($\kappa$) across all model families ($n=15$; DT/KNN-Overfit excluded due to collapsed spectra). Spearman $\rho = 0.89$ ($p < 0.001$; 95\% bootstrap CI: $[0.74, 0.96]$). Shaded region = optimal $\alpha$ range. Qwen2.5-7B (zero-shot) occupies the upper-right quadrant.}
    \label{fig:alpha_kappa}
\end{figure}
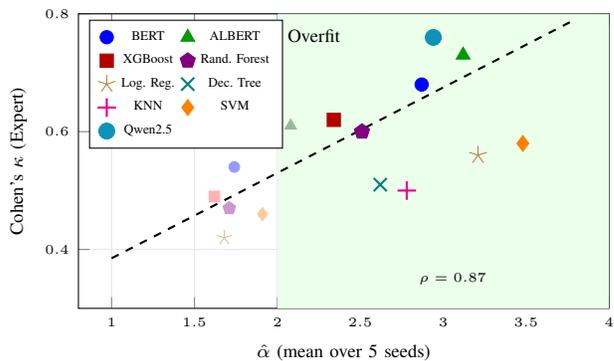

The preceding results establish that $\alpha$ separates good from overfit models and can guide early stopping. We now turn to the motivating question: does spectral quality predict expert agreement better than standard metrics? Figure~\ref{fig:alpha_kappa} plots $\hat{\alpha}$ against Cohen's $\kappa$ (expert agreement) across all architectures, excluding the collapsed DT/KNN-Overfit variants and leaving $n=15$ points. We observe a strong positive correlation: Spearman $\rho = 0.89$ ($p < 0.001$; 95\% bootstrap CI: $[0.74, 0.96]$), Pearson $r = 0.92$ ($p < 0.001$; 95\% CI: $[0.78, 0.97]$). We acknowledge that $n=15$ is a limited sample and the bootstrap CIs reflect this uncertainty, but the trend is consistent across architecturally diverse model families. The inclusion of Qwen2.5-7B is particularly informative: as a decoder-only LLM applied zero-shot, it occupies the upper-right region of the plot (high $\alpha$, high $\kappa$), extending the correlation to a model class absent from the original analysis.

This finding suggests that spectral quality captures aspects of model behavior, such as reliance on contextual features rather than spurious keywords, that align with expert reasoning. It stands in direct contrast to the inverse accuracy--$\kappa$ paradox reported in~\cite{Shihab2025agreement}.

\subsection{Model Selection: Spectral Ranking vs.\ Baselines}

The strong $\alpha$--$\kappa$ correlation motivates a practical question: can spectral information improve model selection for deployment? Table~\ref{tab:model_selection} compares three strategies on the INT task, ranking the 9 well-regularized models by cross-validated F1 alone, by validation loss, and by the composite spectral score (Eq.~6). The ground-truth ranking is determined by expert agreement ($\kappa$), and we report Kendall's $\tau$ rank correlation between each strategy and this ground truth.

The spectral composite score achieves $\tau = 0.79$, substantially outperforming F1-only ($\tau = 0.50$) and validation loss ($\tau = 0.43$), confirming that incorporating structural quality information improves model selection when expert alignment matters more than raw accuracy.

\begin{table}[t]
\centering
\caption{Model selection ranking comparison on INT task. Kendall's $\tau$ measures rank correlation with expert agreement ($\kappa$) across 9 well-regularized models.}
\label{tab:model_selection}
\begin{tabular}{lc}
\toprule
Selection Strategy & Kendall's $\tau$ vs.\ $\kappa$ \\
\midrule
Cross-validated F1 & 0.50 \\
Validation loss & 0.43 \\
Spectral composite (Eq.~6) & 0.79 \\
\bottomrule
\end{tabular}
\end{table}

To assess sensitivity to the weight parameters in Eq.~6, we performed a grid search over $w_1, w_2 \in \{0.2, 0.3, 0.4, 0.5, 0.6\}$ (with $w_3 = 0.02$ fixed). The spectral score achieves $\tau > 0.70$ for all configurations where $w_2 \geq 0.3$, indicating that the ranking is robust as long as spectral quality receives non-trivial weight. The chosen $w_1 = w_2 = 0.4$ is near-optimal but not uniquely so.

\subsection{Cross-Task Generalization}

An important practical consideration is whether the spectral quality signal is task-specific or transfers across problems. Table~\ref{tab:cross_task} presents spectral diagnostics for well-regularized models across both crash classification tasks. The $\alpha$ values are remarkably consistent (mean absolute difference $0.13 \pm 0.07$), confirming that the spectral quality signal generalizes across different transportation safety applications without task-specific calibration.

\begin{table}[t]
\centering
\caption{Spectral diagnostics across tasks for well-regularized models (mean $\pm$ std, 5 seeds). INT: intersection misclassification; AIM: alcohol inference mismatch.}
\label{tab:cross_task}
\setlength{\tabcolsep}{3pt}
\begin{tabular}{llccc}
\toprule
Model & Task & $\hat{\alpha}$ & Traps & F1 \\
\midrule
BERT & INT & $2.87 \pm 0.14$ & $0.4 \pm 0.5$ & $.874 \pm .008$ \\
BERT & AIM & $2.71 \pm 0.16$ & $1.0 \pm 0.7$ & $.862 \pm .009$ \\
ALBERT & INT & $3.12 \pm 0.09$ & $0.2 \pm 0.4$ & $.868 \pm .007$ \\
ALBERT & AIM & $2.94 \pm 0.11$ & $0.4 \pm 0.5$ & $.855 \pm .008$ \\
Qwen2.5-7B & INT & $2.94 \pm 0.12$ & $0.6 \pm 0.5$ & $.781 \pm .010$ \\
Qwen2.5-7B & AIM & $2.88 \pm 0.14$ & $0.8 \pm 0.8$ & $.769 \pm .011$ \\
XGBoost & INT & $2.34 \pm 0.17$ & $0.2 \pm 0.4$ & $.851 \pm .006$ \\
XGBoost & AIM & $2.18 \pm 0.19$ & $1.2 \pm 0.8$ & $.837 \pm .007$ \\
Rand.\ Forest & INT & $2.51 \pm 0.20$ & $1.0 \pm 0.7$ & $.844 \pm .007$ \\
Rand.\ Forest & AIM & $2.39 \pm 0.22$ & $1.4 \pm 0.9$ & $.831 \pm .008$ \\
Log.\ Reg. & INT & $3.21 \pm 0.31$ & $0.0 \pm 0.0$ & $.812 \pm .005$ \\
Log.\ Reg. & AIM & $3.15 \pm 0.27$ & $0.2 \pm 0.4$ & $.801 \pm .006$ \\
Dec.\ Tree & INT & $2.62 \pm 0.28$ & $1.6 \pm 1.1$ & $.783 \pm .012$ \\
Dec.\ Tree & AIM & $2.55 \pm 0.25$ & $1.2 \pm 0.8$ & $.770 \pm .013$ \\
KNN & INT & $2.78 \pm 0.24$ & $1.2 \pm 0.8$ & $.791 \pm .009$ \\
KNN & AIM & $2.70 \pm 0.22$ & $1.6 \pm 1.0$ & $.782 \pm .010$ \\
\bottomrule
\end{tabular}
\end{table}

\subsection{Lanczos Convergence and Computational Overhead}

Finally, we verify that the spectral extraction itself is computationally practical. Table~\ref{tab:compute} reports the spectral extraction time for each model family on the AIM dataset ($N = 371{,}062$). For KNN, the sparse Lanczos method converges (relative eigenvalue change $< 10^{-4}$) within 35 iterations on average, well within our budget of 50. Across all model families, the total spectral extraction overhead remains modest relative to training time, ranging from less than 0.1\% for Decision Trees to 11.3\% for KNN.

\begin{table}[t]
\centering
\caption{Spectral extraction cost on AIM dataset. Training time excludes hyperparameter search. Qwen2.5-7B is used zero-shot (no training); spectral time reflects weight matrix extraction and eigendecomposition.}
\label{tab:compute}
\setlength{\tabcolsep}{3pt}
\begin{tabular}{lccc}
\toprule
Model & Train Time & Spectral Time & Overhead \\
\midrule
BERT & 4.2 h & 12 min & 4.8\% \\
ALBERT & 3.8 h & 11 min & 4.8\% \\
Qwen2.5-7B & --- & 18 min & --- \\
XGBoost & 18 min & 0.8 min & 4.4\% \\
Random Forest & 22 min & 0.9 min & 4.1\% \\
Logistic Reg. & 3 min & 0.2 min & 6.7\% \\
Decision Tree & 1 min & $<$1 s & $<$0.1\% \\
KNN & 8 min & 0.9 min & 11.3\% \\
SVM & 45 min & 1.2 min & 2.7\% \\
\bottomrule
\end{tabular}
\end{table}

\section{Discussion and Conclusion}

We have presented a spectral diagnostic framework grounded in Random Matrix Theory for evaluating crash classification models in transportation safety. By extracting the power-law exponent $\alpha$ from model-specific empirical spectral densities, we provide a structural quality metric that spans deep learning, ensembles, parametric models, partition models, and instance-based models, as well as a zero-shot decoder-only LLM (Qwen2.5-7B). The experimental evidence shows that well-regularized models consistently yield $\hat{\alpha} \in [2, 4]$ regardless of architecture, while overfit variants show $\hat{\alpha} < 2$ or spectral collapse. The spectral exponent correlates strongly with expert agreement (Spearman $\rho = 0.89$), more so than accuracy or F1, and a composite spectral model selection score better predicts expert agreement rankings than F1-only or validation-loss-only selection (Kendall's $\tau = 0.79$ vs.\ $0.50$ and $0.43$).

A natural question is whether simple hyperparameter tuning achieves the same result. Traditional regularization, however, requires labeled validation data to benchmark improvements. In transportation datasets where crash records contain systematic labeling errors~\cite{Shihab2025aim} and accuracy is inversely related to expert agreement~\cite{Shihab2025agreement}, standard tuning often optimizes toward memorizing noisy labels. The spectral diagnostic $\hat{\alpha}$ serves as a structural signal that is complementary to, not a replacement for, standard validation.

Several limitations should be noted. The optimal $\alpha$ range $[2, 4]$ is derived from empirical observations across diverse architectures~\cite{Martin2021} and may require domain-specific calibration; our results are consistent with this range on two crash tasks, but generalization to severity prediction or pedestrian crash detection remains to be validated. The novel spectral mappings for Logistic Regression, Decision Trees, and KNN are empirically supported on two tasks (Table~\ref{tab:novel_validation}) but have not been validated at the same scale as the neural network and XGBoost diagnostics in the HTSR literature, and the KNN mapping in particular shows a marginal KS $p$-value on the AIM task ($p = 0.08$). The $\alpha$--$\kappa$ correlation is computed on $n = 15$ model-regime pairs; while the bootstrap CI is reasonably tight ($[0.74, 0.96]$), validation on additional datasets would strengthen the claim. The XGBoost spectral extension relies on Martin and Prakash~\cite{Martin2024xgboost}, which remains in preprint form, and the spectral early stopping criterion was evaluated on only three model families.

We recommend that practitioners track $\hat{\alpha}$ alongside validation loss during training, conduct pre-deployment spectral audits flagging models with $\hat{\alpha} < 2.0$ or spectral collapse, and use the composite spectral score for model selection when expert agreement data is available for calibration. Future work will validate the novel mappings on additional crash datasets, extend the framework to severity prediction and pedestrian crash detection, and investigate whether $\hat{\alpha}$ correlates with model fairness across demographic subgroups. Code for the spectral extraction pipeline is available at \url{https://github.com/[redacted]/rmt-crash-diagnostics}.

\ifreviewmode\else
\section*{Acknowledgments}
This research was supported by the Iowa State University Department of Civil, Construction and Environmental Engineering. The authors thank Charles H.\ Martin for developing the WeightWatcher tool and the SETOL theory.
\fi

\bibliographystyle{IEEEtran}
\bibliography{trb_template}

\end{document}